# SynForceNet: A Force-Driven Global-Local Latent Representation Framework for Lithium-Ion Battery Fault Diagnosis

RONGXIU CHEN, YUTING SU, LINFENG ZHENG

SINO-German College of Intelligent Manufacturing, Shenzhen Technology University, Shenzhen 518118, China

CORRESPONDING AUTHOR: LINFENG ZHENG (email: zhenglinfeng@sztu.edu.cn)

**Abstract** Online safety fault diagnosis is essential for lithium-ion batteries in electric vehicles(EVs), particularly under complex and rare safety-critical conditions in real-world operation. In this work, we develop an online battery fault diagnosis network based on a deep anomaly detection framework combining kernel one-class classification and minimum-volume estimation. Mechanical constraints and spike-timing-dependent plasticity(STDP)-based dynamic representations are introduced to improve complex fault characterization and enable a more compact normal-state boundary. The proposed method is validated using 8.6 million valid data points collected from 20 EVs. Compared with several advanced baseline methods, it achieves average improvements of 7.59% in TPR, 27.92% in PPV, 18.28% in F1 score, and 23.68% in AUC. In addition, we analyze the spatial separation of fault representations before and after modeling, and further enhance framework robustness by learning the manifold structure in the latent space. The results also suggest the possible presence of shared causal structures across different fault types, highlighting the promise of integrating deep learning with physical constraints and neural dynamics for battery safety diagnosis.

**Keywords**: Lithium-ion batteries, Electric vehicles, Online fault diagnosis, Deep anomaly detection, Deep SVDD, STDP-based dynamic representation, Latent manifold structure

## Introduction

Owing to their high energy density, long cycle life and relatively low maintenance cost, lithium-ion batteries(LiBs) have become the preferred power source for electric vehicles(EVs)[1-5]. However, because of their intrinsic electrochemical mechanisms, the capacity of LiBs inevitably fades with increasing charge-discharge cycles. This electrochemical ageing process not only degrades battery performance, but may also induce internal structural changes and reduced thermal stability, thereby increasing safety risks. In particular, under the complex operating conditions encountered by real-world vehicles, early fault prediction and real-time monitoring of battery failures remain technically challenging.

Current studies on battery anomaly detection can be broadly categorized into four classes.

①Model-driven methods describe battery behaviors by constructing equivalent circuit models or electrochemical models, and identify anomalies by comparing the residuals between model outputs and actual measurements in conjunction with state estimation approaches such as Kalman filtering[6-8]. ② Data-driven methods exploit observable signals, including voltage, current and temperature, to learn the mapping between battery states and fault characteristics through machine learning or deep learning models[9-11]. ③ Knowledge-driven methods develop diagnostic strategies based on expert experience, fault mechanisms or rule bases, such as threshold rules, fault trees and empirical degradation patterns[12-14]. ④ Hybrid-driven methods combine physical models with data-driven approaches, improving the robustness and interpretability of anomaly detection by integrating model residuals, expert knowledge or physical constraints[15-17].

Most existing studies on battery fault diagnosis have been conducted under experimentally simulated conditions. In real-world operating environments, however, fault monitoring usually relies only on the limited signals accessible from the onboard battery management system, while being simultaneously affected by missing and noisy data, inter-vehicle variability and stochastic changes in operating conditions. As a result, diagnostic models developed under laboratory settings are often difficult to transfer directly to real-world scenarios. To advance deployable fault diagnosis for lithium-ion batteries in new energy electric vehicles, there is an urgent need for systematic investigations based on large-scale, high-quality field-operation data covering diverse operating conditions, so as to support the development of robust and generalizable algorithms[18-20]. Using real-world electric vehicle data collected over 18 months, Bin Kaleem et al. developed a fault detection framework that combines segmented regression, gated recurrent units(GRU) and adaptive thresholds to identify voltage abnormalities associated with internal short circuits, external short circuits and capacity fade[21]. Cao et al. utilized 18.2 million valid records from 515 vehicles to propose a model-constrained deep learning network for stochastic operating conditions, achieving online diagnosis of faults including electrolyte leakage, thermal runaway, internal short circuit and abnormal ageing[22]. Yang et al. established a collaborative fault early-warning framework based on 30 real public charging stations, 10,154 vehicles and 21,175 charging sequences, demonstrating that large-scale battery fault warning under highly heterogeneous real-world data conditions has become a feasible engineering pathway[23].

In the field of battery anomaly detection, as battery systems continue to evolve, the distribution patterns of normal samples have become increasingly complex, making it difficult for traditional methods based on fixed thresholds or simple statistical features to accurately characterize their intrinsic structure. Deep SVDD addresses this challenge by learning a compact boundary enclosing normal samples and constructing a tight geometric representation in the latent feature space, thereby enabling sensitive detection of deviations that may indicate incipient faults. Experimental evidence has shown that this approach is particularly effective when anomalous samples are scarce or when the data ecosystem is highly complex[24]. Chan et al. proposed a novel unsupervised SVDD framework that extracts more generalizable features to prevent hypersphere collapse while effectively mitigating ambiguous class boundaries and feature distortion[25]. Liao et al. enhanced the feature extraction capability of SVDD for complex battery data distributions by introducing a scale-learning mechanism[26]. Cheng et al. further improved anomaly recognition under complex data structures and limited fault samples by optimizing the decision boundary in kernel space[27].

Although existing methods have achieved substantial progress in battery anomaly detection, most of them still rely fundamentally on similarity-based metrics as the core of relation modelling. Modern one-class classification techniques and representation learning methods typically use Euclidean distance, Mahalanobis distance, kernel-space inner products, or their variants to quantify the deviation of a sample from the centre of normal patterns, and then establish anomaly decision boundaries accordingly. These approaches offer clear advantages in describing static geometric differences among samples, but their expressive power remains limited when facing the variable coupling, state transitions and dynamic evolutionary relationships that are pervasive in complex battery systems. In practice, the evolution of battery anomalies is not manifested solely as the deviation of a feature point from the normal region at a given moment, but more importantly as changes in the interaction relationships among multiple variables and their cumulative effects over time. Although distance-based measures can characterize the magnitude of deviation, they are insufficient to fully capture the latent interaction mechanisms underlying anomaly formation. Motivated by this understanding, the present study moves beyond a purely distance-constrained framework and further introduces an interaction-modelling perspective inspired by physical dynamics. In this way, inter-sample relationships are extended from a single static description of proximity to a richer representation that simultaneously incorporates directionality, interaction strength and dynamic associations. This provides a potentially more expressive modelling perspective for battery anomaly detection under complex operating conditions, and may help overcome the limitations of traditional distance-based formulations.

Building upon the above considerations, we designed an improved SVDD network that integrates spiking neural encoding, disentangled latent representations and local temporal plasticity constraints, termed SynForceNet(synaptic force network), for battery fault diagnosis under real-world complex operating conditions. The development, training and evaluation of the proposed algorithm were all conducted on real-vehicle datasets. Each data frame was sampled at 10-s intervals as model input, ensuring real-time capability and broad practical applicability. The overall framework can be summarized into four major modules: ① multi-source monitoring feature construction; ② global-local disentangled representation; ③SynForce-constrained relational learning and ④anomaly scoring and diagnostic output. Our validation covered all datasets containing complex operating conditions. The proposed method achieved substantial improvements in both fault point identification and fault type diagnosis, outperforming the best-performing classical anomaly detection algorithm with average gains of 7.59% in TPR, 27.92% in PPV, 18.28% in F1 score and 23.68% in AUC. Taken together, SynForceNet overcomes conventional limitations in representing complex faults and modelling latent relationships in lithium-ion batteries, enabling online diagnosis of multiple fault types throughout the full life cycle and across diverse operating conditions. These results highlight both its theoretical significance and its practical value for safety applications of traction batteries.

# Results

## Framework overview

In addition to coping with fluctuations in online monitoring data, the construction of

SynForceNet faces two further key challenges. First, under continuously varying battery states, the model requires representations that are robust to disturbances arising from complex operating conditions and that can stably characterize the boundary between normal patterns and anomalous deviations. To address this issue, we organize the input as temporally structured monitoring features and construct a global-local dual-branch latent encoding architecture, enabling joint representation of battery states from both the perspective of overall distribution characteristics and local evolutionary relationships. Second, model training is challenged by the severe scarcity of negative samples, which makes it difficult to satisfy the dependence of conventional supervised learning methods on large amounts of labelled fault data. To overcome this bottleneck, we divide the modelling objective into two coordinated levels. At the first level, a stable latent-space structure of normal samples is established through global compactness constraints and reconstruction-consistency learning. At the second level, SynForce relational constraints are introduced to further characterize local dynamic associations, thereby enhancing the model's sensitivity to deviations induced by anomalous patterns. This strategy enables the model to be trained primarily on normal operating data, effectively reducing its reliance on the scale of fault samples while improving detection capability under complex fault conditions.

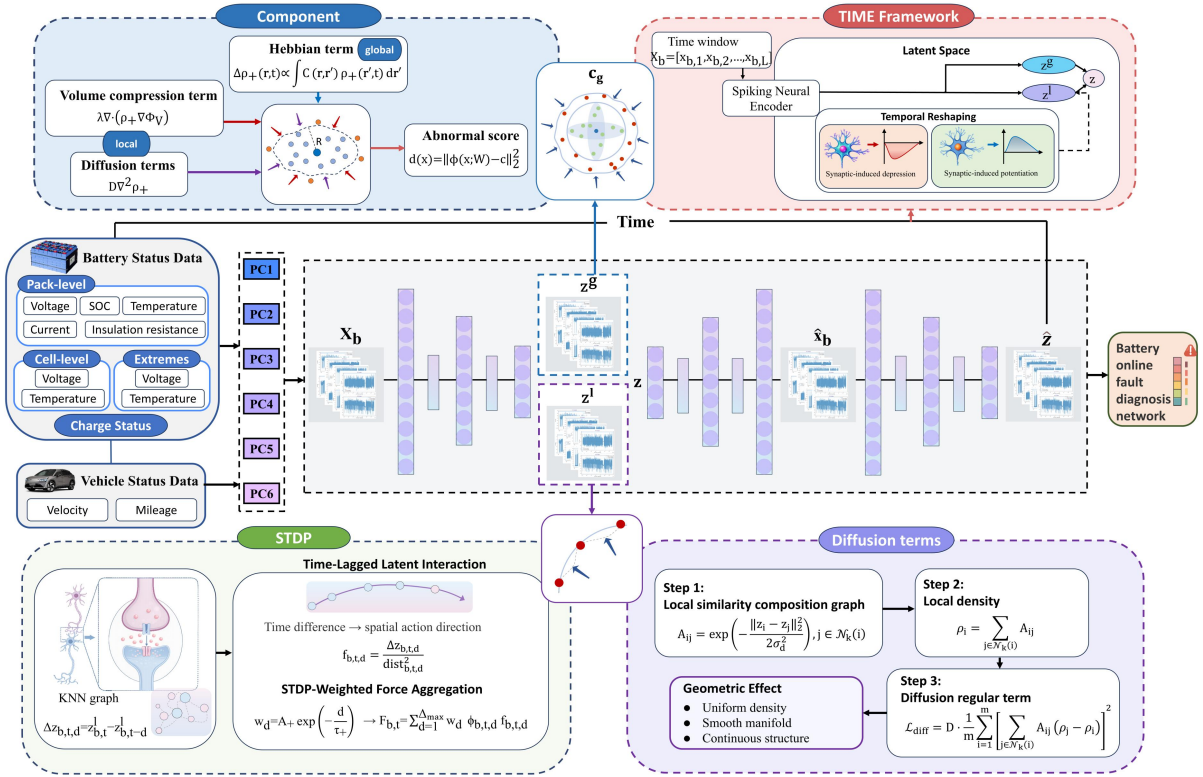

**Fig.1 | SynForceNet Framework.** Component: the global component implemented by the volume compression term, enforces one-class compactness in latent space, while the local component, realized by the Hebbian and diffusion terms, regularizes neighborhood correlation and local manifold smoothness. Their joint effect forms a decoupled global-local latent space for robust anomaly scoring. TIME Framework: a time window $x_b$ is encoded by the spiking neural encoder into two decoupled latent branches, the global latent variable $z^g$, which supports one-class compact representation, and the local latent variable $z^l$, which captures local temporal geometry. The temporal reshaping mechanism acts on $z^l$ to impose synaptic-inspired temporal modulation. Finally, $z^g$ and $z^l$ are fused into the joint latent representation $z=[z^g;z^l]$, which is used as the input to the

decoder. STDP: a KNN graph is first constructed in the local latent space to describe neighborhood relations, and time-lagged latent differences $\Delta z^l_{b,t,d}$ are computed to characterize temporal interactions. These lag-dependent directional terms are then modulated by STDP-inspired exponentially decaying weights and aggregated into a unified local interaction term $F_{b,t}$. This mechanism injects directional temporal bias into the local latent branch, enabling the model to better capture temporally evolving abnormal patterns. Diffusion terms: a local similarity graph is constructed with Gaussian-kernel affinities $A_{ij}$, from which the local density $\rho_i$ is estimated. The diffusion regularizer then encourages neighboring latent points to maintain consistent density, leading to a smoother, more uniform, and more continuous manifold structure. In this way, the diffusion term stabilizes local geometry and improves the separability of abnormal deviations.

**Dataset analysis**

The dataset comprises 8.6 million samples collected from 20 new energy vehicles. After preliminary data preprocessing, it was refined into 2.34 million instances under driving conditions, 1.68 million instances under parking-charging conditions, 0 instances under driving-charging conditions, 6.16 million instances under non-charging conditions, and 40,000 instances under charge-completed conditions. The dataset contains only 3,204 anomalous samples, accounting for approximately 0.037% of the entire dataset. Such an extremely limited proportion poses substantial challenges for the design of risk assessment and fault early-warning algorithms. In the experimental investigation, eight data tables were selected as the study objects, collectively covering 95% of all anomalous samples. In addition, to evaluate the capability of the system to handle multiple battery monitoring tasks under a more limited data regime, we constructed a dataset termed ALL. This dataset contains all anomalous samples together with an equal number of randomly selected normal samples. Across all datasets, the training sets contain only normal samples and are used to learn the boundary of normal behaviour. In contrast, the test sets include all anomalous samples together with an equal number of randomly sampled normal instances, thereby ensuring a balanced ratio of positive and negative examples.

The dataset contains both numerical and categorical features. Specifically, it includes seven groups of categorical features, namely CHARGE STATUS, MAX VOLT CELL ID, MIN VOLT CELL ID, MAX TEMP PROBE ID, MIN TEMP PROBE ID, MAX ALARM LEVEL, and COMMON ALARM FLAG. It also contains 137 groups of numerical features, including TIME. Using the BMS monitoring data collected during online vehicle operation as the raw input, we constructed a multi-level feature system consisting of three categories of information: central features describing the overall operating status of the vehicle and battery pack, cell-voltage features characterizing cell-level differences and local distribution properties, and temperature-probe features reflecting the spatial distribution of thermal states. The central features include charging status, vehicle speed, cumulative mileage, total voltage, total current, state of charge(SOC), insulation resistance, and variables related to voltage and temperature extrema; the cell-voltage and temperature-probe features are, respectively, composed of individual cell-voltage channels and temperature-sampling channels. For each of the three feature categories, robust denoising, normalization, variance weighting and PCA-based dimensionality reduction were performed separately. The principal-component representations preserving the major variance information were then concatenated to form a unified low-dimensional input feature vector, which was finally organized into fixed-length temporal windows as the model input sequence.

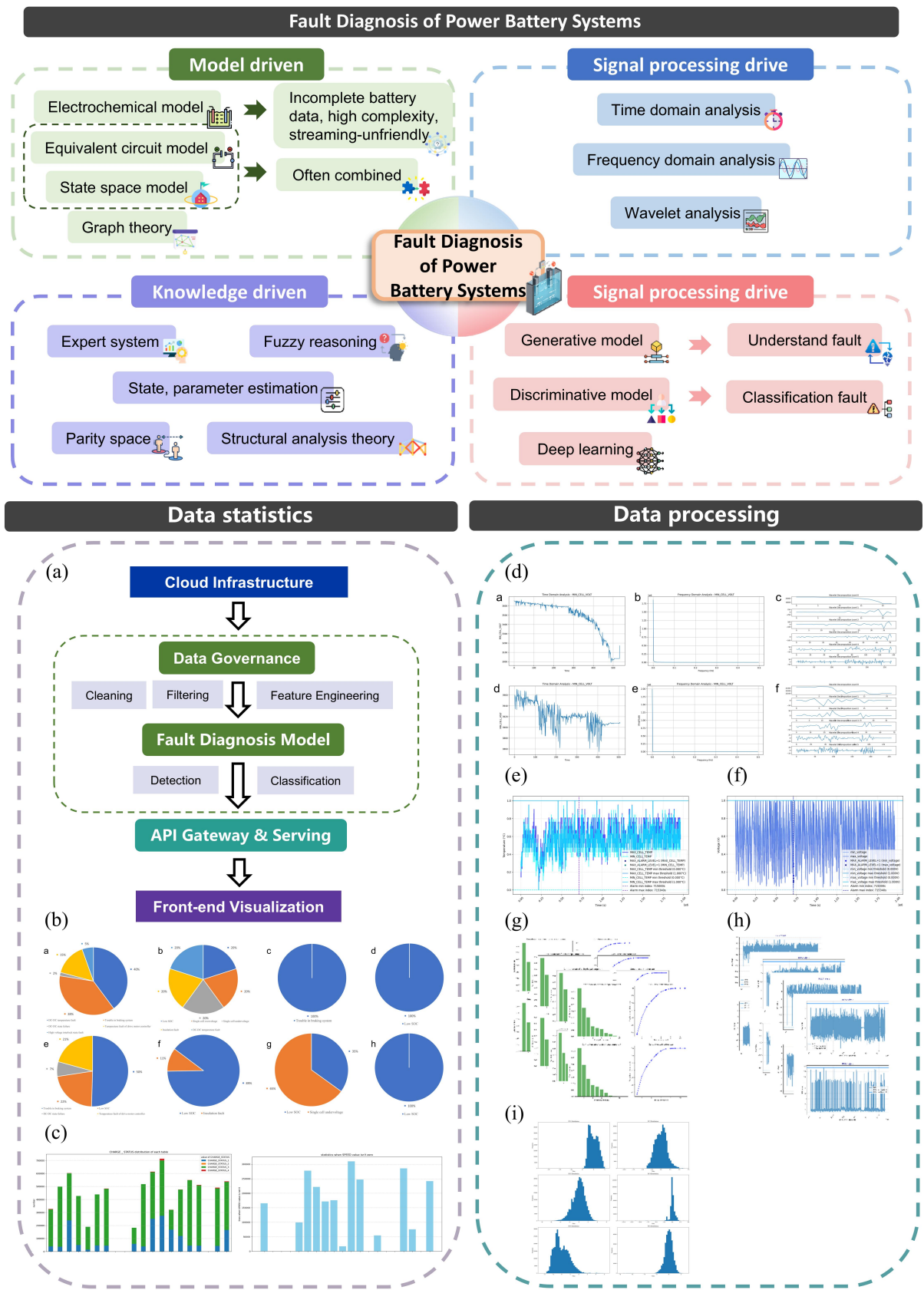

**Fig.2| Fault Diagnosis of Power Battery Systems:** Overview of four categories of battery fault diagnosis strategies and their representative methods.
**Data statistics:** (a) Cloud-based EV data pipeline. Data streams transmitted through the cloud infrastructure undergo data governance procedures, including cleaning, filtering, and feature engineering. The processed data are then fed into the fault diagnosis model for detection and classification, with the outputs delivered to the API gateway and serving layer for front-end visualization. (b) Distribution of fault types across eight subsets (a-h), obtained by partitioning 95% of the full dataset according to the actual test time. (C) Left, distribution of charge status for each dataset. Right, statistics of samples with nonzero speed . Dataset names are abbreviated because of repeated prefixes. (d) Time-frequency analysis of min_cell_volt from one representative dataset. (e,f) Normalized voltage and temperature, respectively, from the same dataset. (g-i) Representative analyses from four datasets, including the contribution rate and cumulative contribution rate of principal components, temporal analysis of principal components, and feature histograms.

**Fault diagnosis**

Under real-world operating conditions, battery degradation often exhibits diverse evolutionary patterns, and the mechanisms that trigger it are highly uncertain. Consequently, during real-time monitoring, most faulty samples do not manifest as clearly distinguishable outliers in the data representation. Figures 3a-d and 3A-D present two-dimensional and three-dimensional visualizations of the distributions of the complete samples and faulty samples in the low-dimensional space. The results show that different fault types, as well as their compound forms,exhibit distinct clustering structures and boundary ranges in the projected space. Meanwhile, the three feature groups-center, U_edge, and T_edge-also display markedly different coverage relationships with respect to faulty samples. In Figs. 3a-d, there is no obvious separation between faulty and normal samples, and multiple fault types coexist within the same regions of the feature space, making faulty samples difficult to distinguish. By contrast, in Figs. 3e-h, faulty samples are distributed mainly along the boundaries of the mapped feature-space distribution,making them easier to identify. These observations highlight the clear performance advantage of the proposed network in addressing the challenges presented by the datasets in Figs. 3a-d.

Specifically, in the two real-world cases shown in Figs. 3a, 3A, 3d, and 3D, the lithium batteries involve a relatively large number of fault clusters and a large total number of fault samples. Most samples belonging to the same fault type tend to follow a certain common distribution,although a small portion of the same-type faults still deviate from the main distribution and form another pattern. In addition, some envelope-like structures and interactions can be observed among different fault distributions, which lays the groundwork for our subsequent investigation into whether different faults may share certain common-cause structures that can be represented in a specific feature space. In contrast, in the two real-world cases shown in Figs.3b, 3B, 3c, and 3C, although the fault categories are relatively single, the fault samples are extremely sparse and scattered, without an obvious clustering trend, which also poses challenges for online diagnosis.

SynForceNet enables online diagnosis of battery safety faults by generating point-wise anomaly scores for each monitoring time step, and the temporal evolution of these scores can further characterize the changing trend of battery health status. Figures 4a-h present the vehicle identification results for nine types of safety faults. During fault identification, three alarm thresholds are defined according to the degree of deviation from the model. Once the level-1alarm threshold is triggered, the fault probability at the alarm point is further calculated. To evaluate the representational differences of different fault patterns across multiple feature groups, the proportional statistics of each feature group further indicate that different faults do not share the same sensitivity to global statistical features, voltage-edge features, and temperature-edge features. This suggests that a single-feature perspective is insufficient to fully characterize complex fault patterns, whereas the multi-level joint feature network constructed in this study helps improve fault representation capability and identification robustness

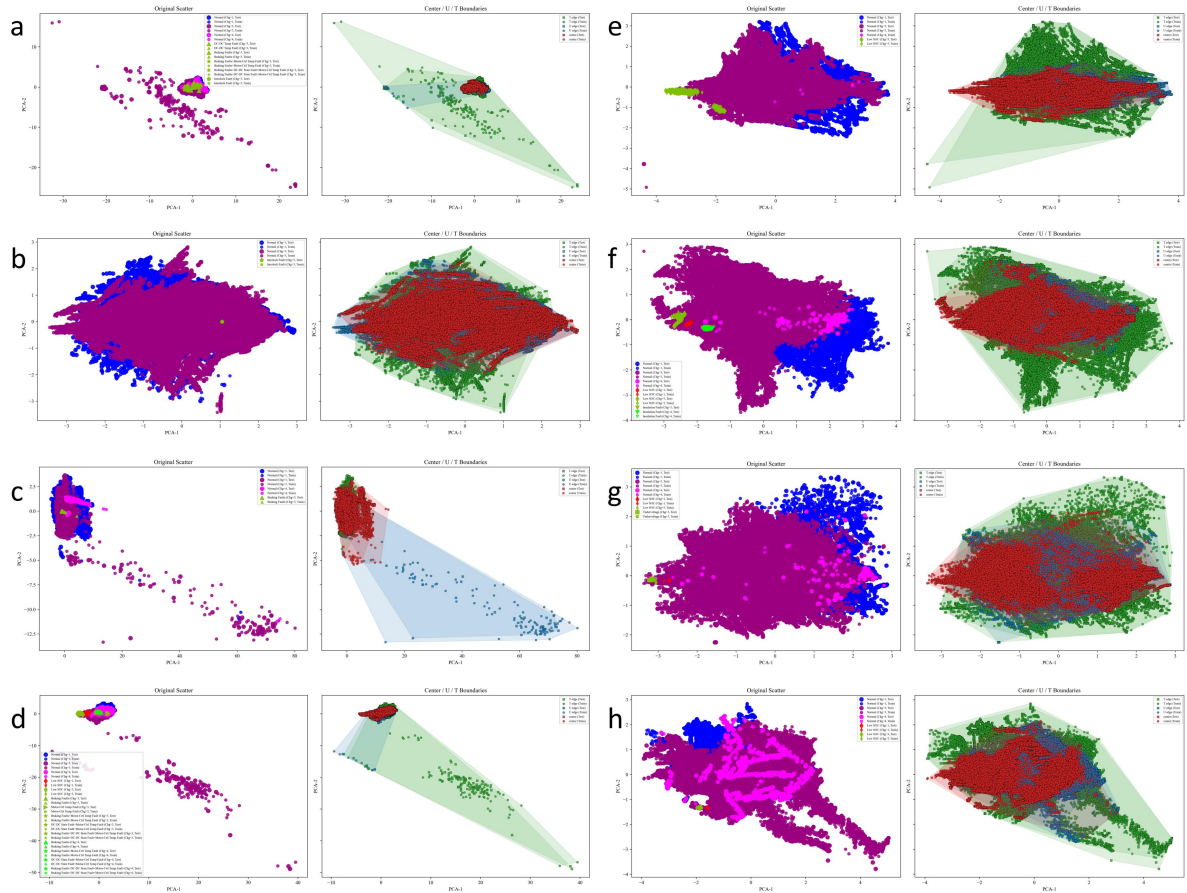

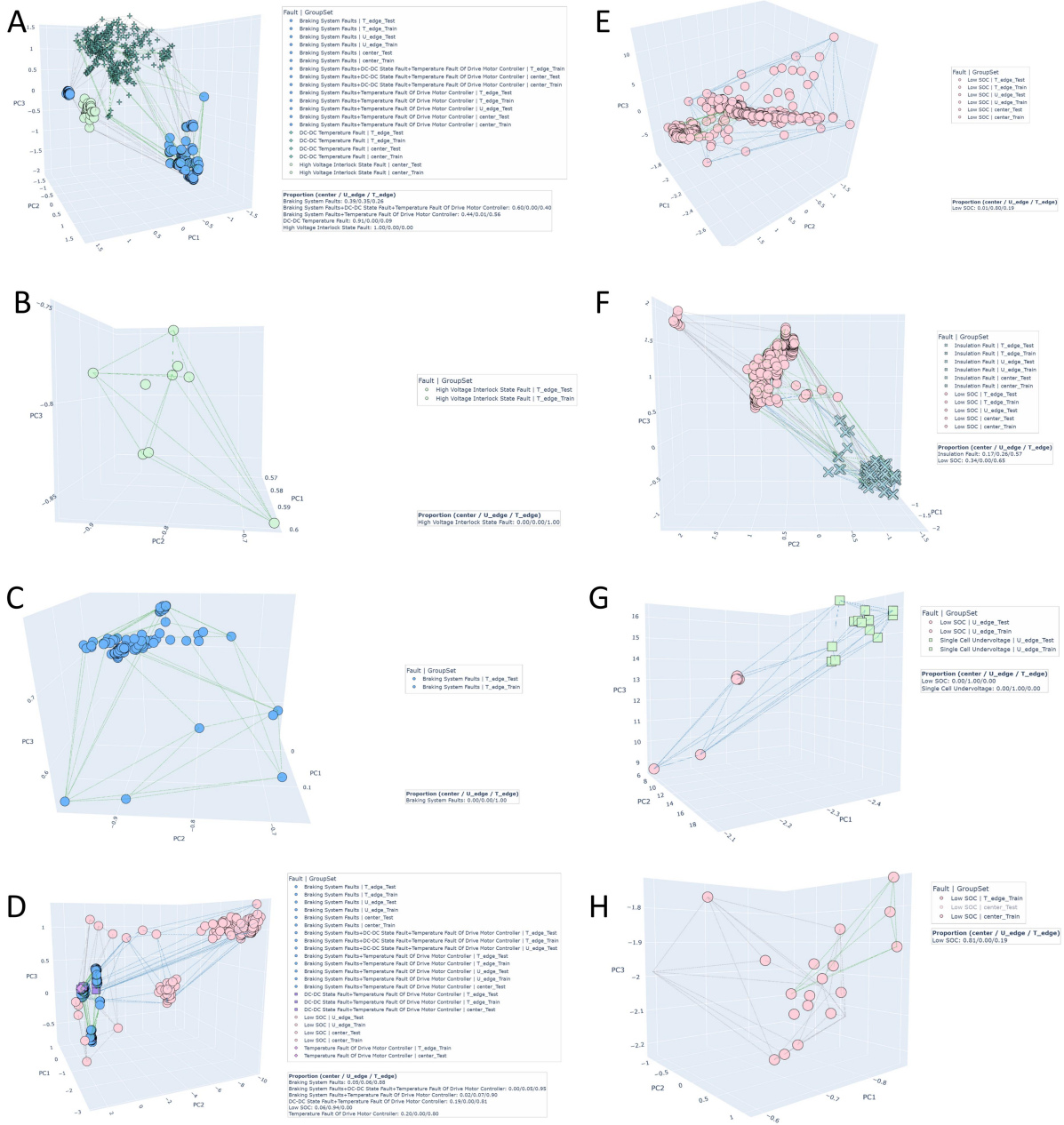

**Fig.3|Two-dimensional and three-dimensional visualizations of the distributions of the complete samples and faulty samples in the low-dimensional space**

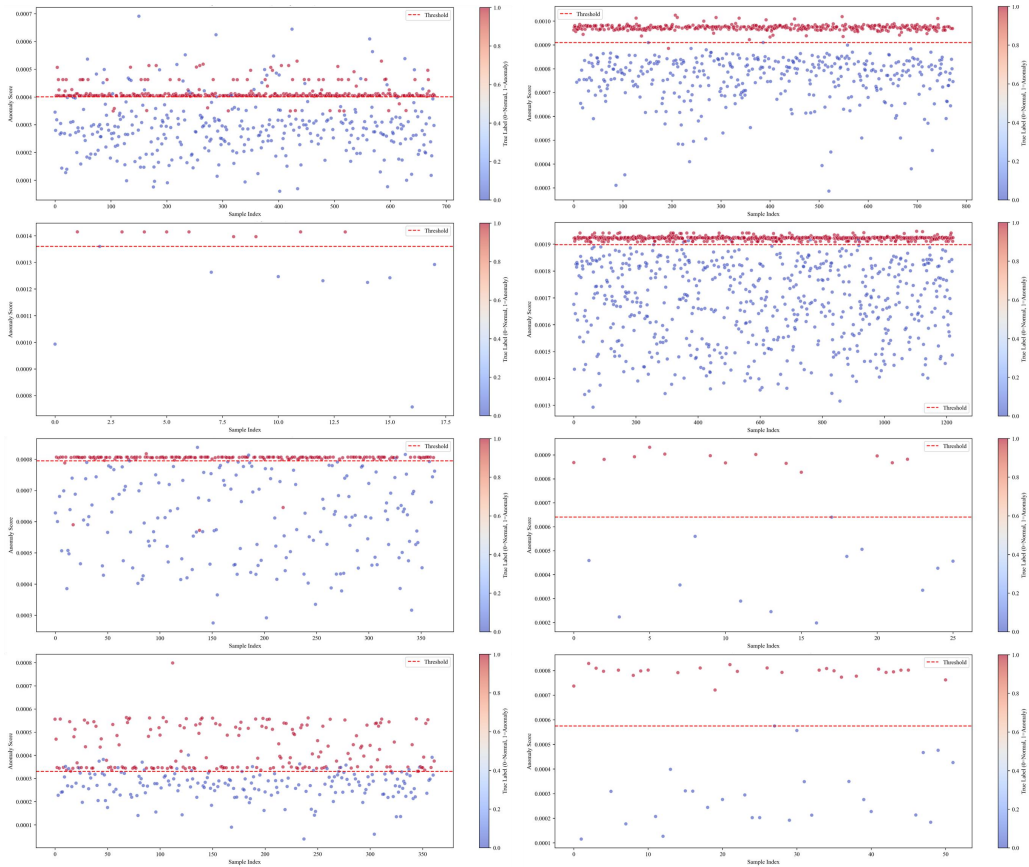

**Fig.4|Vehicle identification results for nine types of safety faults**

**Performance testing**

To validate the advancement of the proposed algorithm, we compared it with representative baseline methods on the validation datasets in terms of both battery-pack status and cell-level status. First, we compared the overall performance of four widely used machine-learning-based fault diagnosis algorithms on the test samples. These methods include Variational Autoencoder(VAE), Sparse Autoencoder(SAE), Deep SVDD, and DVAA-SVDD. The results show that, owing to the high false-positive rates encountered in fault detection under full operating conditions, existing methods often struggle to meet the requirements of practical applications. By contrast, the proposed method achieves an average improvement of 7.59% in TPR, 27.92% in PPV, 18.28% in F1 score, and 23.68% in AUC over the compared methods,demonstrating its substantial performance advantage and stronger applicability. Taken together, SynForceNet exhibits clear superiority in application-oriented battery-pack/cell-state anomaly detection as well as ROC-related performance. This favourable performance can be attributed to the collaborative global-local latent-space modelling strategy, the introduction of the SynForce relational constraint mechanism, and the joint optimization of reconstruction consistency and geometric-structure regularization.

# Discussion

SynForceNet maintains the compactness and stability of normal-sample representations while more effectively distinguishing hard-to-identify faults and complex operating conditions in the latent space. Unlike conventional approaches that rely solely on the static distance between samples and the hypersphere centre, SynForceNet further emphasizes local temporal relationships within the latent space, inter-sample interactions, and the dynamic organization of the normal manifold. In doing so, it alleviates issues such as blurred boundaries, distorted latent representations, and unstable anomaly characterization. Specifically, the network first employs a spiking-neural-inspired encoder to learn representations from the input temporal windows and decomposes the latent representation into a global component and a local component. The global component is used to establish a compact centre-oriented description of normal samples, thereby maintaining the overall geometric boundary required for anomaly detection; the local component, by contrast, is used to characterize short-term dynamic variations and is reshaped through an STDP mechanism with stabilized directional constraints, so that latent trajectories evolve more consistently along the temporal axis. Meanwhile, reconstruction-consistency constraints, a diffusion regularizer and a volume-compression term are introduced during training to jointly suppress latent-space collapse and excessive local discreteness, thereby improving the structural stability of the normal manifold and the separability of anomalous samples. During testing, the model generates anomaly scores solely from the distance between the global latent component and the centre, thus preserving the simplicity of the detection criterion; the local latent component and its STDP-based dynamical constraints mainly act during training to shape the representation, thereby enhancing the deviation tendency of hard-to-identify faults in the latent space.

By combining latent-space distribution visualization with geometric-structure analysis, we provide a structured interpretation of how SynForceNet improves fault-detection performance. The results show that the representations of anomalous samples in the latent space undergo an evolutionary process from entanglement and overlap to outward boundary displacement and local aggregation, indicating that the network enhances the geometric separability between normal and anomalous samples. We further argue that this representational optimization does not arise solely from boundary contraction itself, but is also related to the model's ability to capture shared driving factors of anomalies. Although different fault types may differ in their external manifestations, their latent representations often jointly reflect disruption of normal coupling relationships, changes in dynamic interaction patterns, and increased local instability. By introducing an interaction-inspired relational modelling mechanism, SynForceNet more effectively captures these anomaly factors shared across fault types, thereby driving anomalous samples from within or near the normal distribution boundary towards regions outside the boundary and forming clearer local aggregation structures.

# Methods

### ①Force-driven latent-space modelling

The core objective of Deep SVDD is to progressively compress the distribution volume of normal samples in the latent space during network training, such that normal samples are compactly clustered around a centre, whereas anomalous samples lie outside the hypersphere. The original formulation and its corresponding optimization objective can be expressed as follows:

$$\mathscr{F}=\{z\in\mathbb{R}^p \mid \|z-c\|_2^2 \leq R^2\}, \quad \min_{\mathbf{W},R} \frac{1}{n}\sum_{i=1}^{n} \|\phi(x_i;\mathbf{W})-c\|_2^2 + \lambda\|\mathbf{W}\|_2^2.$$

Here, $c$ denotes the centre of the hypersphere, and $(\phi(\cdot;\mathbf{W}))$ denotes the deep feature mapping. During the testing stage, for any sample $x$, its anomaly score is typically defined as $d(x)=\|\phi(x;\mathbf{W})-c\|_2^2$. However, characterizing the relationships among samples, as well as between samples and the hypersphere centre, solely in terms of distance or similarity remains essentially a static geometric description. Such a formulation is unable to explicitly capture directionality, multimodal interaction channels, temporal derivatives, or latent causal relationships. Particularly in battery time-series anomaly detection, relying only on the distance between a sample and the centre neglects the interaction structure among normal samples, among anomalous samples, and between normal and anomalous samples, as well as the dynamical constraints governing the evolution of the latent space. To address these limitations, we attempt to extend the latent-space paradigm from a purely distance-based formulation to a force-driven formulation. Based on this idea, we map the normal sample distribution, the hypersphere centre, and anomalous samples in the latent space to a positive-charge-like cloud constrained within the hypersphere, a negative-charge-like source, and free charges, respectively, thereby constructing a physically interpretable unified-field-like framework. Let the hypersphere centre be $\mathbf{r}_0 \in \mathbb{R}^N$, and let the continuous distribution density of normal samples in the latent space be denoted by $\rho_+(\mathbf{r},t)$, where $\mathbf{r}\in B^N(R)=\{\mathbf{r}\in\mathbb{R}^N \mid \|\mathbf{r}-\mathbf{r}_0\|_2 \leq R\}$. Anomalous samples are represented as discrete particles $(q_k, \mathbf{r}_k(t))$. Under this framework, the dynamics of the normal sample cloud reflect not only local cooperative activation, but also global volume compression and density smoothing mechanisms.

According to the Hebbian learning principle that co-activation strengthens connectivity[28-29], the local correlation can be written in the following kernel-integral form:

$$\Delta\rho_+(\mathbf{r},t) \propto \int C(\mathbf{r},\mathbf{r}')\rho_+(\mathbf{r}',t)\,d\mathbf{r}'.$$

Here, $C(\mathbf{r},\mathbf{r}')$ is chosen as a Gaussian kernel,

$$C(\mathbf{r},\mathbf{r}')=\exp\left(-\frac{\|\mathbf{r}-\mathbf{r}'\|_2^2}{2\sigma^2}\right),$$

which describes the spatial decay of local cooperative activation with distance. To minimize the latent volume occupied by normal samples and suppress distributions far from the hypersphere centre, we define the volume potential energy as

$$\Phi_V = \frac{1}{V(B^N)} \int_{B^N} \|\mathbf{r}-\mathbf{r}_0\|_2^3 \, \rho_+(\mathbf{r},t)\, d\mathbf{r},$$

where $V(B^N)$ denotes the volume of the $N$-dimensional hypersphere, and the cubic term $\|\mathbf{r}-\mathbf{r}_0\|_2^3$ is introduced to impose a stronger penalty on distributions located far from the centre. Driven by the

gradient of this potential energy, density contraction further yields the volume-compression term

$$-\lambda \nabla \cdot (\rho_+ \nabla \Phi_V).$$

To prevent unbounded density aggregation and preserve smoothness inside the latent space, we further introduce a diffusion mechanism:

$$\frac{\partial \rho_+}{\partial t} = D \nabla^2 \rho_+,$$

where $D$ is the diffusion coefficient. Taken together, the continuous dynamics of the normal sample cloud in the latent space can be unified as

$$\frac{\partial \rho_+}{\partial t} = \underbrace{\eta \int_{B^N} \rho_+(\mathbf{r}',t) C(\mathbf{r},\mathbf{r}') \, d\mathbf{r}'}_{\text{Hebbian term}} - \underbrace{\lambda \nabla \cdot (\rho_+ \nabla \Phi_V)}_{\text{volume-compression term}} + \underbrace{D \nabla^2 \rho_+}_{\text{diffusion term}}.$$

Nevertheless, directly solving the above unified-field-like continuous model during deep network training faces three major difficulties. First, $\rho_+(\mathbf{r},t)$ is a continuous density in a high-dimensional latent space, and its explicit estimation depends on high-dimensional density modelling, spherical-domain integration, and partial differential equation solving, which are difficult to implement stably under a mini-batch stochastic gradient descent framework[30-32]. Second, the current anomaly detection task remains essentially a one-class learning problem, where only normal data are used during training; therefore, anomalous particles $(q_k, \mathbf{r}_k)$ cannot directly participate in the dynamical update in the same way as in supervised learning[33-35]. Meanwhile, coupling centre-oriented compression, anomaly repulsion, and STDP modulation with positive and negative signs within a unified latent space essentially corresponds to multiple potentially competing optimization objectives. Such joint constraints are prone to introducing gradient conflicts and instability in representation learning. In one-class anomaly detection, instability in latent representations may further propagate into bias in the normality boundary, fluctuations in the centre position, and increased threshold sensitivity, ultimately weakening the consistency of anomaly scoring[36-38]. In light of these challenges, while retaining the three core ideas of volume compression, local diffusion, and temporal-causal driving, the present study reformulates the original theoretical conception into a trainable form.

②**Global-local disentangled latent-variable network (SynForceNet)**

Specifically, rather than directly solving the continuous density evolution equation, we discretize it into a latent representation learning problem induced by a deep encoder. Given a normal window of length $L$,

$$X_b = [x_{b,1}, x_{b,2}, \ldots, x_{b,L}], \quad x_{b,t} \in \mathbb{R}^d,$$

the training set is defined as the collection containing only normal windows:

$$D_{\text{tr}} = \{X_b \mid \sum_{t=1}^{L} y_{b,t} = 0\}.$$

The encoder adopts a spiking-neural-inspired architecture and explicitly decomposes the latent variables into a global component and a local component:

$$z_{b,t}^g = E_g(x_{b,t}), \quad z_{b,t}^l = E_l(x_{b,t}), \quad z_{b,t} = [z_{b,t}^g; z_{b,t}^l].$$

The global component $z^g$ is mainly responsible for one-class description and anomaly scoring, ensuring that normal samples are compactly contracted towards the hypersphere centre. By contrast, the local component $z^l$ is dedicated to encoding local temporal geometric relationships and is used to introduce STDP-style directional constraints. Compared with simultaneously imposing compactness and temporal forces in a single latent space, this decoupled strategy of "global discrimination-local dynamics" is more conducive to alleviating objective conflicts.

It is worth noting that, in the original conceptual design, we considered extending a variational autoencoder(VAE) to generalize normal features so as to form a more coherent latent manifold. However, in one-class anomaly detection, the stochastic sampling and KL regularization introduced by VAEs tend to actively expand the latent distribution[39-40], creating an intrinsic tension with the compact normal-sample enclosure pursued by Deep SVDD and potentially resulting in blurred hypersphere boundaries and threshold drift. Therefore, in the final engineering implementation, we did not adopt an explicit VAE structure. Instead, we employed a deterministic encoder-decoder framework and further introduced an encoding consistency constraint, so as to preserve reconstruction capability while avoiding excessive divergence of latent representations.

Specifically, the decoder reconstructs the input from the joint latent variables:
$$\hat{x}_{b,t}=D(z_{b,t}),$$
and the corresponding reconstruction loss is
$$\mathcal{L}_{\text{rec}} = \frac{1}{BL}\sum_{b=1}^{B}\sum_{t=1}^{L} \|\hat{x}_{b,t} - x_{b,t}\|_2^2.$$

For the one-class description branch, the actual implementation uses the unsquared Euclidean norm. Accordingly, the global SVDD loss is written as
$$\mathcal{L}_{\text{svdd}} = \frac{1}{BL}\sum_{b=1}^{B}\sum_{t=1}^{L} \|z^g_{b,t} - c_g\|_2,$$
where $c_g$ denotes the centre of the global latent space. The use of the unsquared norm reduces the dominance of extreme outliers on the gradient, thereby leading to a more stable centre-oriented compression during training[41].

Considering that the reconstruction results produced by the decoder may in turn alter the latent structure, we further introduce an encoding consistency constraint. Let
$$(\hat{z}^g_{b,t}, \hat{z}^l_{b,t}) = E(\hat{x}_{b,t}), \quad \hat{z}_{b,t} = [\hat{z}^g_{b,t}; \hat{z}^l_{b,t}],$$
then the consistency loss is defined as
$$\mathcal{L}_{\text{enc}} = \frac{1}{BL}\sum_{b=1}^{B}\sum_{t=1}^{L} \|\hat{z}_{b,t} - \text{sg}(z_{b,t})\|_2^2.$$

Here, sg(·) denotes the stop-gradient operation. This term is not directly derived from the original theoretical formulation, but is introduced in the engineering implementation as an additional stabilizing term to ensure compatibility between "reconstruction in the input space" and "compression in the latent space".

③**Physics- and STDP- inspired structural constraints in the latent space**

In the continuous-field model, the diffusion term of $\rho_+(\mathbf{r},t)$ is used to smooth the density distribution in the latent space. In the discrete implementation, we construct a k-nearest-neighbour graph on the mini-batch latent point set $\{z_i\}_{i=1}^{m}$, and define
$$A_{ij} = \exp\left(-\frac{\|z_i-z_j\|_2^2}{2\sigma_d^2}\right), \quad j \in \mathcal{N}_k(i), \quad \rho_i = \sum_{j \in \mathcal{N}_k(i)} A_{ij}.$$

The diffusion regularization term is written as

$$\mathcal{L}_{\text{diff}} = D \cdot \frac{1}{m}\sum_{i=1}^{m} \left[\sum_{j\in\mathcal{N}_k(i)} A_{ij}(\rho_j - \rho_i)\right]^2,$$

which can be viewed as a discrete approximation to the continuous diffusion equation $D\nabla^2\rho$. Its purpose is to prevent local spikes and discontinuous collapse in the latent representation.

Corresponding to the volume potential $\Phi_V$ in the theoretical formulation, we further design a surrogate term for volume compression. Let the joint latent dimension be $p$, and define the extended hypersphere centre as

$$\tilde{c}=[c_g;\mathbf{0}].$$

For an arbitrary latent point $z_i$, we define its radial distance to the extended centre as

$$d_i=\|z_i-\tilde{c}\|_2,$$

and the density surrogate based on the radial distance as

$$\tilde{\rho}_i=\exp\left(-\frac{d_i^2}{2\sigma_v^2}\right).$$

Let the maximum radius within the current batch be

$$R=\max_i d_i,$$

then the corresponding hypersphere volume is approximated by

$$V(B^p(R))=\frac{\pi^{p/2}}{\Gamma\left(\frac{p}{2}+1\right)}R^p.$$

The volume-compression term is defined as

$$\mathcal{L}_{\text{vol}} = \beta \frac{\sum_{i=1}^{m} d_i^3 \tilde{\rho}_i}{\left(\sum_{i=1}^{m}\tilde{\rho}_i\right)V(B^p(R))}.$$

This term does not explicitly solve the gradient flow induced by $\Phi_V$, but instead follows the core idea that points farther from the centre should be penalized more strongly, and that larger occupied volume should incur a greater penalty. It can therefore be regarded as a differentiable surrogate of the original volume-potential model.

On this basis, we further aim to use STDP to reshape the latent space temporally. The central mechanism of STDP is that the change in synaptic weight depends on the relative timing difference between pre- and postsynaptic spikes[42-43]. Its standard temporal window function can be written as

$$A(\Delta t)=A_+\exp\left(-\frac{\Delta t}{\tau_+}\right)-A_-\exp\left(\frac{\Delta t}{\tau_-}\right).$$

When $\Delta t>0$, the presynaptic neuron fires before the postsynaptic neuron, corresponding to long-term potentiation; when $\Delta t<0$, the interaction corresponds to long-term depression. To map this causal directionality into particle interactions in the latent space, we construct a spike-induced driving term:

$$\sum_l A(t_k-t_l)\frac{\mathbf{r}_k-\mathbf{r}_l}{\|\mathbf{r}_k-\mathbf{r}_l\|_2^3}.$$

Meanwhile, anomalous particles are also subject to the combined influence of the hypersphere centre and the normal sample cloud. The equation of motion for a free charge can therefore be written as

$$\frac{d\mathbf{r}_k}{dt} = \sum_l A(t_k-t_l)\frac{\mathbf{r}_k-\mathbf{r}_l}{\|\mathbf{r}_k-\mathbf{r}_l\|_2^3} + \alpha\left(\frac{\mathbf{r}_k-\mathbf{r}_0}{\|\mathbf{r}_k-\mathbf{r}_0\|_2^3} - \int_{B^N}\frac{\rho_+(\mathbf{r},t)(\mathbf{r}_k-\mathbf{r})}{\|\mathbf{r}_k-\mathbf{r}\|_2^3}\,dV\right).$$

Furthermore, if the entire latent space is regarded as a neural field or potential field, the following Poisson-type equation can be written:

$$\nabla^2 \Phi(\mathbf{r}, t) = -\frac{1}{\epsilon}\left[\rho_+(\mathbf{r}, t) - \rho_- \delta(\mathbf{r} - \mathbf{r}_0) + \sum_k q_k \delta(\mathbf{r} - \mathbf{r}_k)\right].$$

The above formulation provides a unified interpretative framework for the interactions among the hypersphere centre, the normal sample cloud and anomalous particles. Its key contribution is to provide a unified dynamical description for four phenomena in the latent space: compression, diffusion, local causal interaction and anomalous deviation.

In the original theoretical conception, the STDP effect was written as a particle interaction term with explicit temporal differences ($t_k - t_l$), simultaneously including both long-term potentiation and long-term depression. In the actual implementation, however, we do not directly use real spike times; instead, we approximate the relative temporal difference using the discrete time lag $d$ within a fixed-length window. At the same time, to avoid oscillations in the latent space caused by the simultaneous action of positive and negative terms, we retain in the engineering implementation only a stabilized one-sided directional constraint, acting exclusively on the local latent variable $z^l$.

Let the local latent sequence of the b-th sample window be

$$\{z^l_{b,1}, z^l_{b,2}, \ldots, z^l_{b,L}\},$$

and define the local motion vector at time $t$ as

$$v_{b,t} = z^l_{b,t+1} - z^l_{b,t}.$$

For each lag $d=1,2,\ldots,\Delta_{\max}$, we first define the difference vector

$$\Delta z_{b,t,d} = z^l_{b,t} - z^l_{b,t-d},$$

and the corresponding softened distance

$$\text{dist}^2_{b,t,d} = \|\Delta z_{b,t,d}\|^2_2 + \varepsilon^2,$$

where $\varepsilon > 0$ is introduced to avoid singularities in the denominator. Based on this, the local force direction term is defined as

$$f_{b,t,d} = \frac{\Delta z_{b,t,d}}{\text{dist}^2_{b,t,d}},$$

and the Gaussian local interaction gate is defined as

$$\phi_{b,t,d} = \exp\left(-\frac{\|\Delta z_{b,t,d}\|^2_2}{2\sigma^2_f}\right).$$

The corresponding temporal decay weight is written as $w_d = A_+ \exp\left(-\frac{d}{\tau_+}\right)$.

The local STDP resultant force can then be approximated by

$$F_{b,t} = \sum_{d=1}^{\Delta_{\max}} w_d \, \phi_{b,t,d} f_{b,t,d}.$$

To align the actual motion direction of the local latent sequence as closely as possible with the STDP-induced direction, we do not directly minimize the Euclidean distance between them. Instead, we minimize the cosine loss of the angle between the two[44]. Let

$$\bar{v}_{b,t} = \frac{v_{b,t}}{\|v_{b,t}\|_2 + \epsilon}, \qquad \bar{F}_{b,t} = \frac{F_{b,t}}{\|F_{b,t}\|_2 + \epsilon},$$

then the local STDP regularization term is defined as

$$\mathcal{L}_{\text{stdp}} = \frac{1}{B(L-\Delta_{\max}-1)}\sum_{b,t}\left(1 - \bar{v}_{b,t}^{\top}\bar{F}_{b,t}\right).$$

This design indicates that the STDP component in the actual implementation is no longer a strict biological spike-based synaptic update equation, but rather a local dynamical regularizer that takes temporal decay as a prior and directional consistency as the optimization target. The purpose is not to reproduce all potentiation/depression mechanisms of real neural systems, but to embed a temporally structured bias with causal directionality into the latent space[45].

In addition, considering that $\mathcal{L}_{\text{rec}}$, $\mathcal{L}_{\text{svdd}}$ and $\mathcal{L}_{\text{enc}}$ differ substantially in numerical scale, we do not directly combine them using fixed linear weights. Instead, we introduce learnable uncertainty parameters ($s_r, s_s, s_e$), and the final total loss is written as

$$\mathcal{L} = \left(e^{-2s_r}\mathcal{L}_{\text{rec}} + s_r\right) + \left(e^{-2s_s}\mathcal{L}_{\text{svdd}} + s_s\right) + \left(e^{-2s_e}\mathcal{L}_{\text{enc}} + s_e\right) + \omega_{\text{diff}}\mathcal{L}_{\text{diff}} + \omega_{\text{vol}}\mathcal{L}_{\text{vol}} + \lambda_{\text{stdp}}\mathcal{L}_{\text{stdp}}.$$

Here, $\omega_{\text{diff}}$, $\omega_{\text{vol}}$ and $\lambda_{\text{stdp}}$ control the strengths of diffusion regularization, volume compression and STDP directional constraints, respectively.

### ④Training strategy, centre update and anomaly scoring

For centre updating, we adopt a smoothed epoch-wise update strategy. Let $\bar{c}_g^{(e)}$ denote the global latent mean obtained from all normal training samples at the current epoch. The centre is then updated as

$$c_g^{(e+1)} = 0.99\, c_g^{(e)} + 0.01\, \bar{c}_g^{(e)}.$$

The purpose of this strategy is to alleviate the centre mismatch caused by the rapid variation of the encoder during the early stage of training, thereby improving the consistency between the centre and the manifold of normal samples.

The final anomaly score is computed solely from the global latent component:

$$s(x) = \|E_g(x) - c_g\|_2.$$

This means that the local STDP-related latent component is intentionally not used directly for anomaly scoring; instead, it serves during training as a mechanism for shaping local geometric plasticity in the latent space. Such a design preserves the simplicity and interpretability of the scoring function, while also avoiding direct interference of local dynamic information with the stable estimation of the hypersphere boundary.

During validation, we construct a point-level balanced evaluation set with a 1:1 ratio between normal and anomalous samples, and select the threshold corresponding to the optimal F1 score from the Precision-Recall curve, so as to assess the overall performance of the model in separating normal and anomalous samples.

*Algorithm: Procedure of SynForceNet*

1) $Number\ of\ epochs = 50$
2) $Window\ length = 60, training\ stride = 5$
3) $Batch\ size = 64, learning\ rate = 10^{-4})$
4) $Latent\ dimension = 6, where)D_{zg} = 2\ and)D_{zl} = 4$
5) $Total\ loss: L_{total}$
$$= (e^{-2s_r}L_{rec} + s_r) + (e^{-2s_s}L_{svdd} + s_s) + (e^{-2s_e}L_{enc} + s_e)$$
$$+ \omega_{diff}L_{diff} + \omega_{vol}L_{vol} + \lambda_{stdp}L_{stdp}$$

6) $\omega_{diff} = 0.01, \omega_{vol} = 0.01, \lambda_{stdp} = 0.0125$
7) $A^+ = 1.0, \tau^+ = 5.0$
8) $MAX\_LAG = 6, \sigma_{force} = 2.0, \varepsilon = 0.05$
9) $Balanced\ evaluation\ strategy: random$

## Data availability

The data that support the findings of this study are not publicly available due to privacy and confidentiality restrictions, but may be available from the corresponding author upon reasonable request and with permission from the data provider.

## Code availability

The code used to support the findings of this study is available in a GitHub repository at:https://github.com/jinxuanchern/battery.